\documentclass[graybox]{svmult}

\newcommand\numberthis{\addtocounter{equation}{1}\tag{\theequation}}

\usepackage{mathptmx}       
\usepackage{helvet}         
\usepackage{courier}        
\usepackage{type1cm}        
\usepackage{makeidx}         
\usepackage{graphicx}        
\usepackage{amssymb}
\usepackage{amsmath}
\DeclareMathOperator*{\argmax}{arg\,max}
\DeclareMathOperator*{\argmin}{arg\,min}
\usepackage{multicol}        
\usepackage[bottom]{footmisc}
            
\usepackage{url}

\usepackage[numbers]{natbib}

\usepackage{cancel}

\makeindex             

\begin{document}

\title*{Towards Precise Robotic Grasping by Probabilistic Post-grasp Displacement Estimation}
\author{Jialiang (Alan) Zhao \and Jacky Liang \and Oliver Kroemer}
\institute{Jialiang (Alan) Zhao \and Jacky Liang \and Oliver Kroemer \at Robotics Institute, Carnegie Mellon University, 5000 Forbes Avenue, Pittsburgh, PA 15213,\\
\email{{alanjz, jackyliang, okroemer}@cmu.edu}}
%
%
\maketitle

\abstract{
Precise robotic grasping is important for many industrial applications, such as assembly and palletizing, where the location of the object needs to be controlled and known. However, achieving precise grasps is challenging due to noise in sensing and control, as well as unknown object properties.
We propose a method to plan robotic grasps that are both robust and precise by training two convolutional neural networks - one to predict the robustness of a grasp and another to predict a distribution of post-grasp object displacements.
Our networks are trained with depth images in simulation on a dataset of over $1000$ industrial parts and were successfully deployed on a real robot without having to be further fine-tuned.
The proposed displacement estimator achieves a mean prediction errors of $0.68$cm and $3.42$deg on novel objects in real world experiments. 
}
\section{Introduction}
\label{sec:1}

Grasping is one of the most fundamental skills for robots performing manipulation tasks.
Grasping allows a robot to gain control over objects and subsequently use them to perform specific interactions or use them as tools. 
Recent work on grasping has largely focused on the problem of getting the object in the hand in some manner.
However, many manipulation tasks will require the object to be held in a specific and known pose.
For example, to insert a peg in a hole, the robot should apply a firm grasp at the far end of the peg from the insertion point. 
The robot will also need to have an estimate of the object’s pose relative to the hand to perform the actual insertion task. 
Some grasps will allow the robot to constrain the object more than others, reducing the variance in the object’s pose, and thus allow for easier in-hand localization.

\begin{figure}[!ht]
    \centering
    \includegraphics[width=\linewidth]{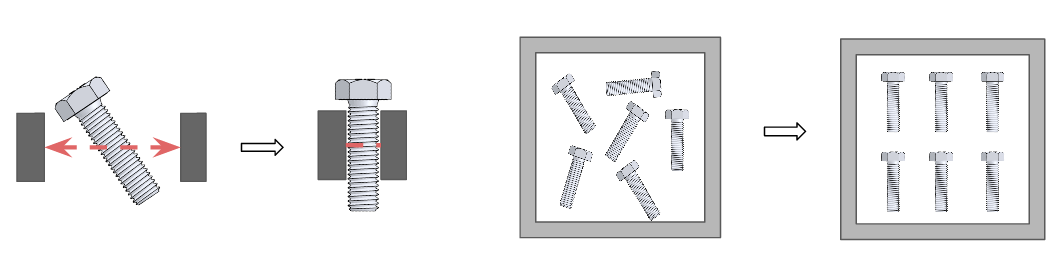}
    \caption{Example of post-grasp object displacement that our method predicts (left) and palletizing as an example application that requires precise grasp planning (right).}
    \label{fig:palletizing}
\end{figure}

In this paper, we address the problem of predicting object displacements during grasping. 
We propose a method that uses two neural networks - the first predicts whether a grasping action will result in a successful grasp that allows an object to be lifted, and the second predicts a distribution over post-grasp object displacements.
The robot then selects grasps with a high success probability and low object displacement variance. 
The predicted mean object displacement can then be used to estimate the in-hand pose for downstream tasks such as assembling and palletizing objects (see Figure~\ref{fig:palletizing}).

Although our system is trained only in simulation, we were able to successfully deploy the networks on a real Franka Panda robot for precise grasping of 3D printed industrial parts without further fine tuning. Videos, datasets, and supplementary material are available at: \url{https://precise-grasping.jialiangz.me}.

\section{Related Works}
\label{sec:2}

To generalize grasps between objects, many recent works have used data-driven grasp synthesis~\cite{bohg2014data} techniques to predict high quality grasps from vision observations.
While early works used hand-tuned visual features~\cite{saxena2008robotic}, recent methods have focused on using Convolutional Neural Networks (CNNs) to learn grasp quality functions from a large amount of training data~\cite{mahler2017dex, mahler2019learning, zeng2018robotic}.

Robots can collect training data for grasping in a self-supervised manner by attempting thousands of random grasps and observing whether or not they result in successful lifts~\cite{kalashnikov2018qt, levine2018learning, pinto2016supersizing}.
By contrast, collecting grasping data in simulation can be much faster and less costly. 
However, learning on simulation data often suffers from the simulation-to-reality (sim2real) gap, where the visual appearance and object dynamics of simulated data deviate from their real world counterparts.
Methods for overcoming the sim2real gap include domain randomization~\cite{tobin2018domain} and domain adaptation~\cite{bousmalis2017unsupervised, james2018sim}.
For grasping, the sim2real gap can be reduced by using only depth images for the visual input.
Depth-only grasping methods used in~\cite{detry2017task, gualtieri2018learning, mahler2017dex, mahler2019learning} do not require further fine-tuning or domain adaptation to perform well in the real world.

Recent works have extended data-driven grasp synthesis for task-oriented grasping, where the system plans grasps that optimize for the success of downstream tasks.
\citet{detry2017task} had human experts label parts of objects that are suitable for tasks like handover and pouring. They trained a CNN to segment depth images based on task suitability to guide the grasp selection.
Other works forgo human labels and use simulations and self-supervision to jointly optimize grasp selection and policies for downstream tasks such as sweeping and hammering~\cite{fang2018learning} or tossing~\cite{zeng2019tossingbot}.

Due to noise in sensing and actuation, as well as unknown object properties, the grasp that the robot intends to execute is often not the grasp that is actually achieved.
\citet{jentoft2018think} analyzed sources of grasp variations and quantified basins of attraction for multi-fingered grasps.
\citet{dogar2011framework} learned from human strategy and used pushing to funnel the clutter of objects before grasping to reduce uncertainty.
\citet{gupta2018robot} explicitly learned a noise model to compensate for actuation noise of low-cost robot arms; while this method improves grasp success, it does not explicitly optimize for precise grasps.
\citet{chen2018probabilistic} combine a probabilistic signed distance function representation of object surfaces from depth images with analytical grasp metrics to plan grasps that optimize for small post-grasp object displacements.

Instead of choosing precise grasps that minimize object displacement, other works have explored estimating the in-hand object poses using additional sensory signals, e.g., vision and tactile~\cite{chebotar2014learning}.
To address the challenge of in-hand occlusions, \citet{choi2016using} trained a CNN to segment out robot grippers such that localization can be done on only the object-relevant pixels, and \citet{izatt2017tracking} used tactile sensing to provide point cloud data on the occluded parts of grasped objects.

In this work, we address the challenges of optimizing grasps for tasks that require precise post-grasp object poses.
Given this context, we note that it is acceptable for a grasp to result in a significant object displacement as long as the robot can reliably predict the displacement and adapt the task execution accordingly.
Our approach stands in contrast with previous works as we train a CNN to predict the expected post-grasp object displacement and the variance of the displacement.
In this manner, the grasp planner can choose grasps that are robust and have the lowest displacement variance.
\section{Learning Precise Robotic Grasping}
\label{sec:3}

In this section we describe the problem of precise grasping and explain our approach for addressing this problem.

\subsection{Problem Statement}
Our proposed approach addresses the problem of estimating a distribution of post-grasp object displacements of top-down, parallel-jaw grasps of singulated objects lying on a flat surface.
The method needs to generalize over novel objects unseen during training.
We are motivated by the palletizing application and therefore focus our experiments on rigid objects commonly found in industrial and manufacturing settings, such as gears, brackets, and screws.

Let the initial pose of an object on the work surface be $\mathbf{p}$.
We assume that during manipulation, the object can only undergo translational movements $(\Delta x,\Delta y,\Delta z)$ and planar rotation $\Delta \theta$, and we define the post-grasp object displacement as $\Delta\mathbf{p}=[\Delta x, \Delta y, \Delta z, \Delta \theta]^T$.
As the robot will not be able to use additional sensors to perform in-hand localization, it needs to predict the displacement $\Delta\tilde{\mathbf{p}}$ based on the object pose $\mathbf{p}$ and observation $\mathbf{o}$. The observation $\mathbf{o}\in\mathbb{R}^{64\times 64}$ is a depth image of the object. We use either object-centric full images or grasp-centric image patches of the object for the observations. In both cases the image size is $64\times 64$ pixels.

A grasp $\mathbf{g}$ has 4 degrees of freedom $\mathbf{g} = [g_x, g_y, g_z, g_\theta]^T$.
The position parameters $(g_x, g_y, g_z)\in \mathbb{R}^3$ denote the location of the grasp relative to the object's geometric center $\mathbf{p}$.
The orientation parameter $g_\theta \in [-\pi, \pi)$ denotes the planar rotation of the gripper about an axis orthogonal to the table surface.

Rather than predicting the post-grasp object displacement $\Delta\mathbf{p}$ directly, our networks instead predict the post-grasp \textit{grasp} displacement $\Delta\mathbf{g}$, i.e., the difference between the grasp parameters and the realized grasp pose relative to the object coordinate frame, which is located at the object's  geometric center. 
This grasp displacement is then converted back to the object displacement $\Delta\mathbf{p}$ for reporting the results.

\subsection{Overview of Approach}
The proposed approach consists of three parts: 1) predicting the grasp quality, 2) predicting the distribution of post-grasp displacements, and 3) combining these predictions to choose robust and precise grasps.

\textbf{Grasp Quality Prediction} 
Following the notation in~\cite{fang2018learning}, we define the grasp quality $Q$ of grasp $\mathbf{g}$ with observation $\mathbf{o}$ as the probability of a successful lift $S$ using the grasp $Q(\mathbf{g}, \mathbf{o}) = P(S = 1 | \mathbf{g}, \mathbf{o})$. We learn this mapping $Q(\mathbf{g}, \mathbf{o})$ using a neural network that we refer to as the Grasp Quality Network (GQN).


\textbf{Grasp Displacement Prediction}
Let $\Delta\mathbf{g}$ denote the distribution of the post-grasp displacement in the object frame. 
We assume that the displacement follows a Gaussian distribution:
\[
\Delta\mathbf{g} \sim \mathcal{N} (\mathbf{\mu}(\mathbf{g}, \mathbf{o}), \mathbf{\sigma}^2(\mathbf{g}, \mathbf{o}))\numberthis
\]
with mean $\mathbf{\mu}(\mathbf{g}, \mathbf{o}) = \{\mu_x, \mu_y, \mu_z, \mu_\theta\}$ and variance $\mathbf{\sigma}^2(\mathbf{g}, \mathbf{o}) = \{\sigma^2_x, \sigma^2_y, \sigma^2_z, \sigma^2_\theta\}$.
We learn a neural network to predict the mean and variance of $\Delta\mathbf{g}$, which we refer to as the Grasp Displacement Network (GDN). 
This network allows the robot to reason about the stochasticity of the grasps and select grasps that minimize the variance over the resulting object poses.

\textbf{Precise Grasp Planning}
Given the two learned networks, we can form a grasp planner that chooses grasps with high probability of successfully lifting the object as well as low variance over the resulting object pose.

\vspace{-2em}

\subsection{Simulation Data}

To generate grasping data for training the networks and generalizing between different objects, we collected $1011$ CAD models of industrial parts such as gears, screws, bolts, and hinges from an online hardware shop\footnote{McMaster-Carr, \url{https://www.mcmaster.com}}. 
Then, in simulation, 1000 random grasp attempts per object were generated and simulated. 
We gathered simulation data using the robotic simulation framework V-REP~\cite{rohmer2013v} with Bullet ver $2.78$~\footnote{\url{https://pybullet.org/}} as the physics engine. 
We assume each object has uniform density.

\begin{figure}[!ht]
    \centering
    \includegraphics[width=\textwidth]{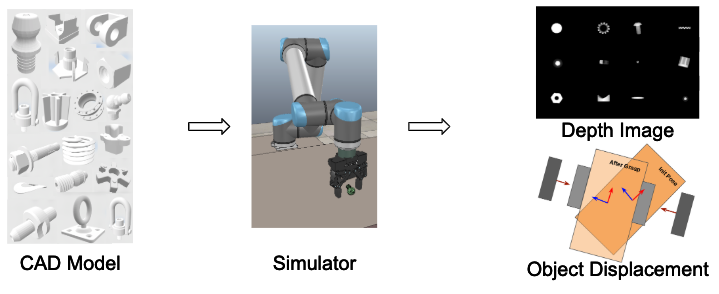}
    \centering
    \caption{\textbf{Simulation Data Collection.} We collected a dataset of $1011$ industrial parts (left). During simulation all objects are set to have uniform density and the same coefficient of friction. We uniformly sampled top-down grasps across the object and evaluate whether or not the grasps resulted in successful lifts (middle). In addition to lift success, we also record the top-down depth image that are used as inputs to our grasp quality and grasp displacement networks (top right). We also recorded the post-grasp object displacement (bottom right).}
    \label{fig:data_collection}
\end{figure}

As shown in Figure~\ref{fig:data_collection}, a depth image is captured at the start of each grasp attempt. 
The robot robot executes a grasp $\mathbf{g}$, where the grasp center $(g_x,g_y,g_z)$ is uniformly sampled from the bounding box of the object, and the grasp orientation $g_\theta$ is uniformly sampled from $[-\frac{\pi}{2}, \frac{\pi}{2})$.
The bounding box's coordinate frame is always parallel to the camera coordinate frame. 
The bounding box is calculated as the minimum area that encloses the entire object in the depth image.
In contrast to previous works that sample antipodal grasps~\cite{mahler2017dex, mahler2019learning}, we use uniformly sampled grasps as using antipodal grasps may introduce a bias in the dataset when estimating post-grasp displacements.
For each grasp attempt $\mathbf{g}$, we record the lift success of the grasp $S$, the overhead depth image $\mathbf{o}$, and the post-grasp object displacement $\Delta \mathbf{p}$.

For training the networks we collected a simulation dataset with $1.011$ million grasps.
After removing objects that are either 1) too hard to grasp (random grasp success rate $<5\%$), 2) too easy to grasp (random grasp success rate $>40\%$), 3) too big (longest axis longer than $15$cm), or 4) too small (longest axis smaller than $2$cm), we have $773$k grasp attempts for $773$ objects. Data is split object-wise, with $660$ objects used for training and $113$ for validation.

\subsection{Grasp Quality and Post-grasp Displacement Estimation}

\begin{figure}[!ht]
    \centering
    \includegraphics[width=\textwidth]{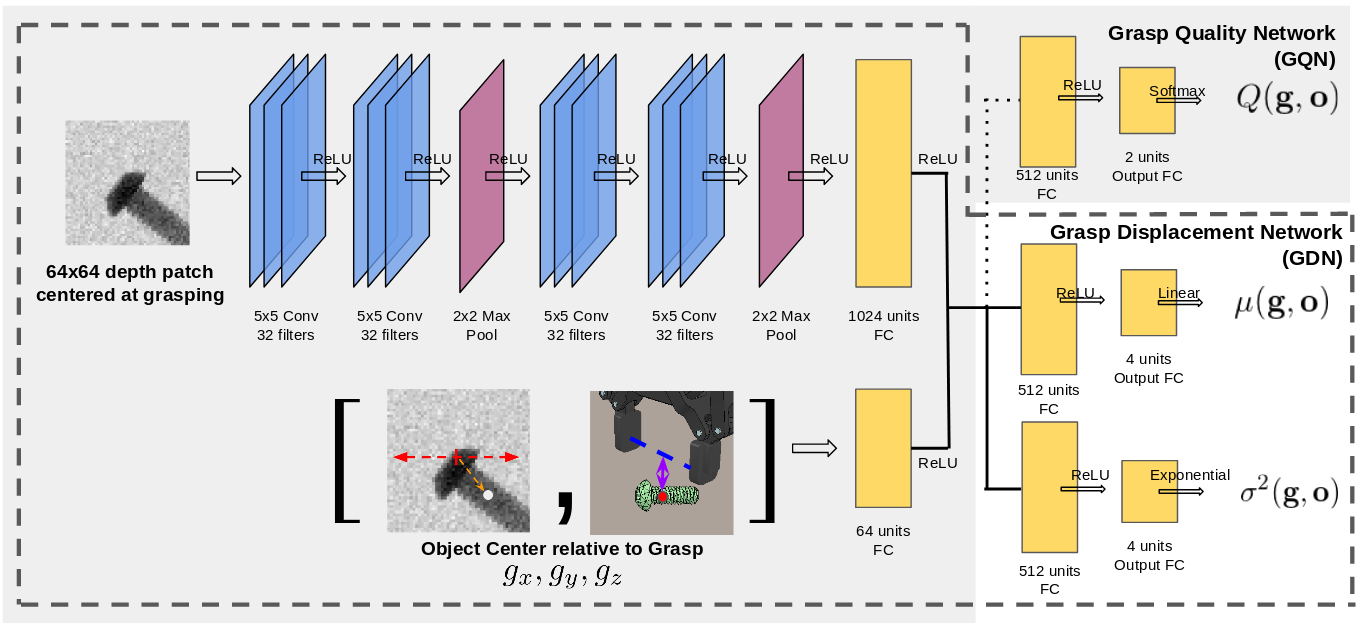}
    \centering
    \caption{\textbf{Grasp Quality Network (GQN) and Grasp Displacement Network (GDN)}. Input to both networks contains a depth image $\mathbf{o}$ cropped at the desired grasp center $\{\mathbf{g_x}, \mathbf{g_y}\}$ aligned to then grasp rotation $\mathbf{g_\theta}$, and the relative translation of the object's geometric center with respect to the grasp center $(g_x, g_y, g_z)$. The GQN predicts grasp quality $Q(\mathbf{g}, \mathbf{o})$, and the GDN jointly predicts the mean $\mathbf{\mu}(\mathbf{g}, \mathbf{o})$ and variance $\mathbf{\sigma}^2(\mathbf{g}, \mathbf{o})$ of post-grasp displacements. The GQN is trained first, and its learned convolution weights are used to initialize the convolution filters of the GDN. The two networks share the same convolution architecture but not the same weights. We use dropout of $0.5$ for the fully connected layers.}
    \label{fig:network}
\end{figure}

We train two types of CNNs - the GQN and the GDN. While the GQN and the GDN share the same convolution architecture, they do not share weights.
Rather, the GQN is trained first, and we use its learned convolution filters to initialize the filter weights of the GDN.
See Figure~\ref{fig:network} for details.

The \textbf{Grasp Quality Network} $Q(\mathbf{g}, \mathbf{o})$ was trained using grasp-centric image patches for the observations, and the translation between the grasp and the object's center $(g_x,g_y,g_z)$. 
We train the GQN using a binary cross entropy loss.


The \textbf{Grasp Displacement Networks} are trained to predict a Gaussian displacement distribution with mean $\mu(\mathbf{g}, \mathbf{o})$ and variance $\sigma^2(\mathbf{g}, \mathbf{o})$. 
For the observations $\mathbf{o}$ we have GDN variants that use the object-centric full images or the cropped grasp-centric image patches. 
The grasp-centric image variants are initialized using the weights from the GQN, while the object-centric image variants are initialized randomly. 
We also have GDN variants that predict only the mean $\mu(\mathbf{g}, \mathbf{o})$ or both the mean $\mu(\mathbf{g}, \mathbf{o})$ and the variance $\sigma^2(\mathbf{g}, \mathbf{o})$. 
These four GDN variants are all trained only on successful $S=1$ grasps and we do not try to predict how failed grasps will displace the objects.


To train the GDN, we form a loss to maximize the log likelihood on the predicted distribution of post-grasp displacement:
\begin{align}
    P(\Delta\mathbf{g} | \mathbf{g}, \mathbf{o}) &= \frac{
        \exp\left(-\frac{1}{2}
        (\Delta\mathbf{g} - \mathbf{\mu}(\mathbf{g}, \mathbf{o}))^\top
        \Sigma(\mathbf{g}, \mathbf{o})^{-1}
        (\Delta\mathbf{g} - \mathbf{\mu}(\mathbf{g}, \mathbf{o}))
        \right)
    }{
        \sqrt{(2\pi)^4 det(\Sigma(\mathbf{g}, \mathbf{o}))}
    } 
     \\
    \mathbf{\mu}^*, \mathbf{\sigma}^{2^*} &= \argmax_{\mathbf{\mu}, \mathbf{\sigma}^2}\ \log(P(\Delta\mathbf{g} | \mathbf{g}, \mathbf{o}))\\
    &= \argmin_{\mathbf{\mu}_j, \mathbf{\sigma}_j^2} \left( \sum_{j=1}^4 \frac{1}{2}\log(\mathbf{\sigma}_j^2) + \frac{(\Delta\mathbf{g}_j - \mathbf{\mu}_j)^2}{2\mathbf{\sigma}_j^2}\right)
\end{align}
Where $\Sigma(\mathbf{g}, \mathbf{o}) = diag(\mathbf{\sigma}^2(\mathbf{g}, \mathbf{o}))$. 
Thus the loss function to train the GDN is:
\begin{equation}
    \label{eq:loss}
    L = \sum_{i=1}^N \left( \sum_{j=1}^4 \log(\sigma^2(\mathbf{g}^{(i)}_j, \mathbf{o}^{(i)}_j)) + \frac{(\Delta\mathbf{g}^{(i)}_j - \mu(\mathbf{g}^{(i)}_j, \mathbf{o}^{(i)}_j))^2}{\sigma^2(\mathbf{g}^{(i)}_j, \mathbf{o}^{(i)}_j)}\right)
\end{equation}

The \textbf{Grasp Planner} needs to select a grasp based on the learned networks.
An ideal grasp for industrial applications needs to satisfy two requirements: (1) the grasp should be stable and (2) the uncertainty over the object's post-grasp pose should be small. 
We fulfill these two requirements by first running the GQN to select the top \textbf{3\%} of the scored randomly generated grasps, denoted as $\mathbf{G}$.
This step makes sure the planned grasp satisfies requirement (1). 
We then choose the optimal grasp $\mathbf{g}^*$ as the one that has the lowest displacement variance predicted by the GDN: $\mathbf{g}^* = \argmin_{\mathbf{g}\in \mathbf{G}} \sigma^2(\mathbf{g}, \mathbf{o})$.
In other words, among all the top successful grasps $\mathbf{G}$, we select the one with the smallest displacement variance.
The expected displacement $\mu(\mathbf{g}^*, \mathbf{o})$ is then used as a correction of the object pose to parameterize downstream tasks such as assembly and palletizing. 
\section{Experiments}
\label{sec:5}

We perform experiments in simulation and the real world to evaluate the performance of both the GQN and the GDNs. 
We also evaluate how selecting low-variance grasps affects performance. The GQN and GDN networks are always trained with simulation data only.

\subsection{Evaluated Grasp Displacement Models}
We compare $5$ models for post-grasp displacement estimation:
\begin{itemize}
    \item \textbf{LOWESS} - Locally Weighted Regression
    \item \textbf{OCFI-M} - GDN with Object-Centric Full Image Input that predicts only the mean grasp displacement.
    \item \textbf{OCFI-M+V} - GDN with Object-Centric Full Image Input that predicts both the mean and the variance of grasp displacement.
    \item \textbf{GCIP-M} - GDN with Grasp-Centric Image Patches that predicts only the mean grasp displacement.
    \item \textbf{GCIP-M+V} - GDN with Grasp-Centric Image Patches that predicts both the mean and the variance of grasp displacement.
\end{itemize}

\textbf{LOWESS}~\cite{cleveland1979robust} is a non-parametric regression method that is similar to nearest neighbors except the weight of the neighbors is computed from a Gaussian kernel.
Because this model doesn't generalize to novel objects, we only use \textbf{LOWESS} as a baseline to predict grasp displacements for \emph{known} objects.
Given $N$ grasps on the known training object, and a new query grasp $\mathbf{g}_j$, the predicted grasp displacement is computed by:
\[
\Delta \hat{\mathbf{g}_j} = \frac{\sum_i^N w(\mathbf{g}_i, \mathbf{g}_j)\Delta\mathbf{g}_i}{\sum_i^N w(\mathbf{g}_i, \mathbf{g}_j)}\numberthis
\]
where $w(\mathbf{g}_i, \mathbf{g}_j) = \mathcal{N}(\mathbf{g}_j|\mathbf{g}_i,\Sigma)$ is the probability density function of the isotropic multivariate Gaussian distribution with mean $\mathbf{g}_i$ and variance $\Sigma$ evaluated at $\mathbf{g}_j$.
We choose the variance terms to be $\Sigma = diag([0.02, 0.02, 0.05, 1.00])$.

The full image GDN variants take as input the uncropped image centered around the object geometric center, instead of the grasp center as is the case with image patches.
These models help us understand how important it is for the GDN to focus on local features around the grasp vs. global features that describe overall object geometry.
Although the GDN variants without variance prediction cannot be used to select grasps by variance, we evaluate against them to see whether or not training to predict this variance helps improve the prediction accuracy of the mean displacements.

\subsection{Training GQN and GDN}
All the CNN models we used have the same convolutional layers structure, the only differences are the input action sizes and output layer activations. 
Before training, all action network inputs are normalized to range $[-1, 1]$, and all depth images are reshaped to $64\times 64$. 
To simulate noise from real depth cameras we add uncorrelated pixel-wise Gaussian noise of zero mean and $3$mm standard deviation to the depth image.
We preprocess all depth images by subtracting their mean and dividing by their standard deviation.

We trained the GQN with balanced positive and negative data for $100$ epochs with the RMSProp optimizer using a learning rate of $10^{-5}$ and decay of $10^{-6}$.
The final accuracy of the GQN is $86.7\%$ on the training set and $85.3\%$ on the validation set. 

\begin{figure}[!ht]
    \centering
    \includegraphics[width=.7\textwidth]{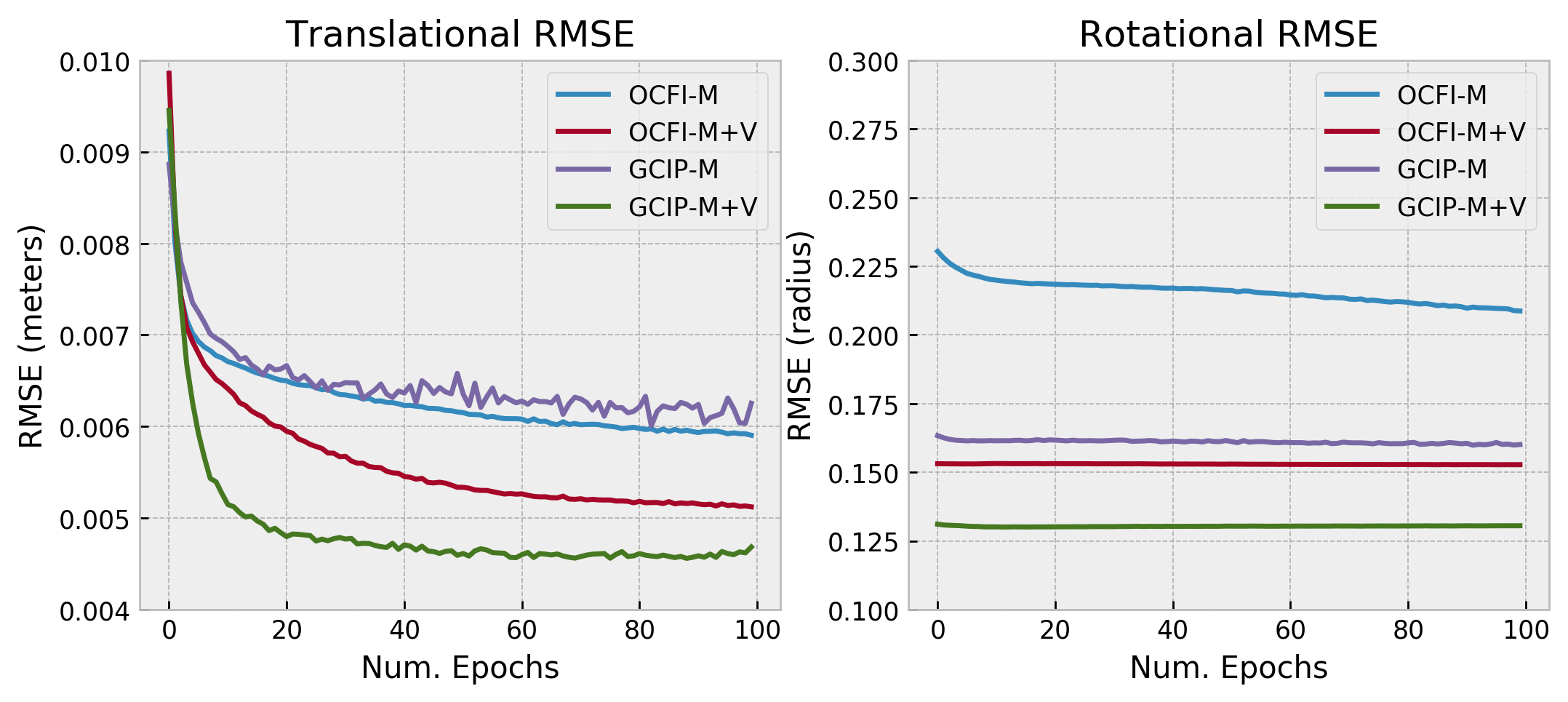}
    \centering
    \caption{\textbf{Validation RMSE of Mean Post-Grasp Object Displacement Predictions for Variants of GDN} We observe that models that incorporate displacement variance prediction consistently outperform ones that do not, and GDN on image patches outperforms operating on the full image.}
    \label{fig:rmse_test}
\end{figure}

We trained all four variants of GDN in a similar fashion.
Root mean square error (RMSE) between predicted mean displacement and actual displacement of each model on the validation set is shown in Figure~\ref{fig:rmse_test}.

We observe that including the displacement variance prediction improves the RMSE for the predicted displacement means.
This improvement is because, with variance prediction, the network is allowed to increase the variance term on data that have high variance instead of fitting a mean which may incur high loss.


\subsection{Evaluation in Simulation}

To evaluate the performance of GQN for planning robust grasps, we formed a grasp planning policy that uniformly samples $3200$ grasps across the object and picks the grasp with the highest predicted quality.
We ran this policy on a set of $130$ novel objects not found in the training or the validation set, and performed $30$ grasp trials for each object.
In simulation the trained GQN achieved a grasp success rate of $\mathbf{94.9}\%$.

To evaluate the performance of the GDNs, we ran two sets of experiments using different grasp selection policies:
(1) select grasps with high predicted qualities according to the GQN and (2) select grasps that have both high quality and low variance. The latter is only applicable for  \textbf{LOWESS}, \textbf{OCFI-M+V}, and \textbf{GCIP-M+V} which predict the variances. 

All of the models in both experiments were evaluated on a set of $85$ objects, of which $50$ are from the training dataset, and $35$ are from the validation dataset.
We perform $35$ grasping trials for each object, and only successful grasps were used to evaluate the performance of displacement prediction models.



\begin{figure}[!ht]
    \centering
    \includegraphics[width=.7\textwidth]{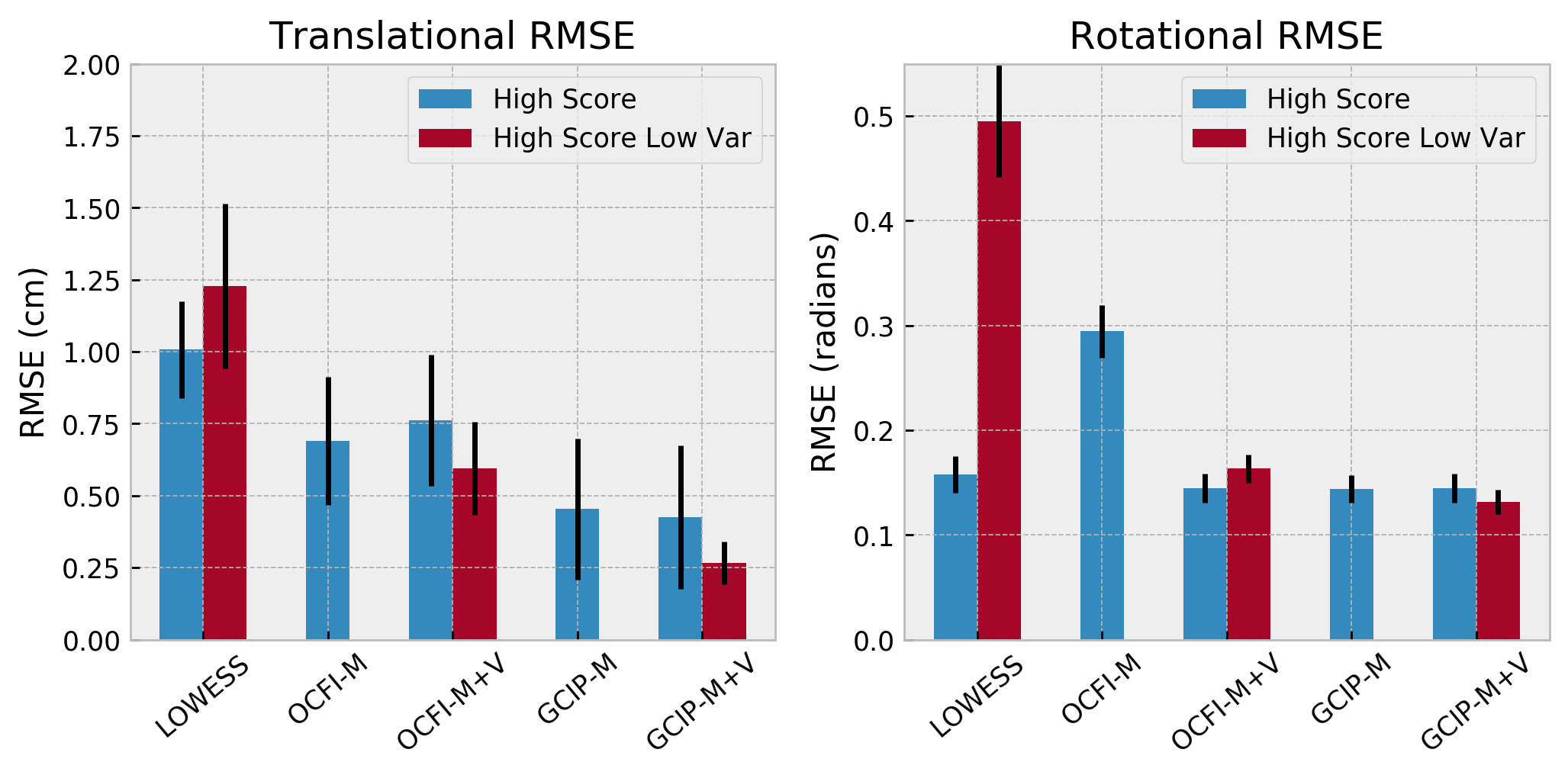}
    \caption{\textbf{Translational and Rotational RMSE of Grasp Displacement Predictions in Simulation.} Blue bars show results of GDN estimated displacement for the grasps with highest qualities predicted by the GQN. Red bars show results of choosing grasps that have both high quality and low variance as predicted by a GDN. Grasp displacement models that do not have variance output are not reported for the second set of experiments.} 
    \label{fig:shift_rmse}
\end{figure}

Results are shown in Fig.~\ref{fig:shift_rmse}. 
For the high-quality grasps, experiment (1), the average translational error and rotational error are $0.43$cm and $8.29$deg with \textbf{GCIP-M+V}. 
For the high-quality low-variance grasps, experiment (2), the errors are further reduced to $0.24$cm and $7.01$deg respectively. 
\textbf{GCIP-M+V} has the best performance in both experiments. 
We also observe that choosing high-quality, low-variance grasps generally improves the post-grasp displacement prediction accuracy.

We observe that in both experiments the proposed \textbf{GCIP-M+V} GDN has a good performance in terms of translational displacement estimation, but it does not reduce the error for rotation in comparison to other baselines.
This might be because predicting translational displacement is easier than rotational displacement, as the latter needs more information regarding the object's overall geometry.

\subsection{Evaluation with Real World Robot}

We use a 7-DoF Franka Emika Panda robot arm for grasping and a Kinect v2 time-of-flight sensor for depth sensing.
For our grasping experiments, we 3D printed $7$ novel objects from our dataset of industrial parts that were not used during training.
At the start of each grasp experiment trial, we place one object in a $24$cm by $24$cm square region in a bin in front of the robot and directly below the depth camera.
Then, the human operator puts a cardboard box over the object before shaking the box for a few seconds.
This helps to reduce human bias and randomize the location and orientation of the object.

During each trial, we use the GQN trained with simulation data to predict the qualities of randomly sampled grasps from the depth image.
A grasp is considered a success if the object is still grasped by the robot after the lift.
If a grasp is successful, the robot proceeds to lowering the end-effector back to the grasp pose before releasing the grippers.

\begin{figure}[!ht]
    \centering
    \includegraphics[width=0.3\textwidth]{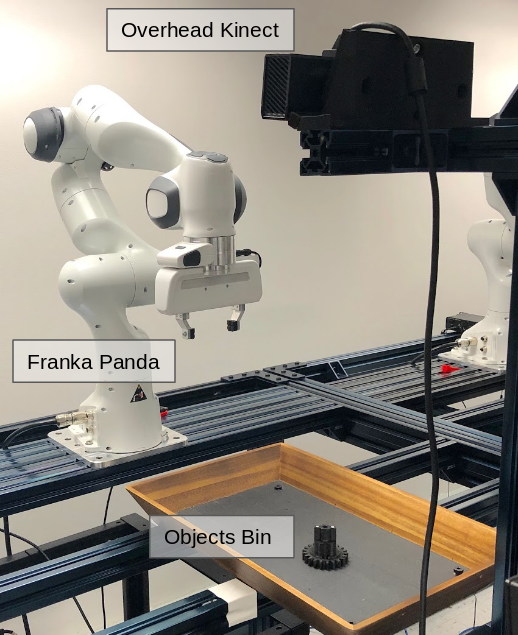}
    \includegraphics[width=0.56\textwidth]{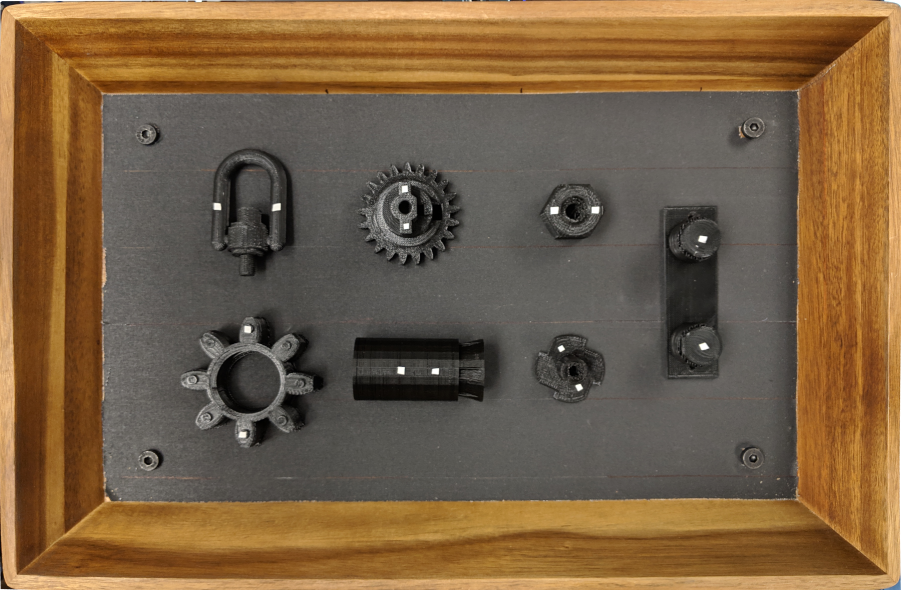}
    \caption{\textbf{Real World Robot Setup.} We use a 7-DoF Franka Panda robot arm and a Kinect v2 RGB-D camera (left) and 3D printed test objects used for real world experiments (right). Two white markers are placed on each object to estimate the post-grasp object displacements.}
    \label{fig:real_world_setup}
\end{figure}

By placing the object at the same gripper pose as the grasping pose, we can compute the relative pre-grasp and post-grasp translation and rotations to estimate object displacement.
One way to do this is via object pose registration with their known 3D CAD models, but this approach is not robust due to rotational symmetries and the low-resolution of the depth sensor.
Instead we opted for a marker-based approach by making the assumption that object displacements during real robot experiments only occur in a plane.
This is done by placing two small, square pieces of white masking tape on top of the object such that the line that connects them crosses the object's geometric center.
Their relative translation and rotation after grasping can be robustly determined from registered color images.
See Figure~\ref{fig:real_world_setup} for robot setup and an illustration of our objects with these markers.


\vspace{-2em}

\subsubsection{Real World Experiment Results}

\begin{figure}[!ht]
    \centering
    \includegraphics[width=.7\linewidth]{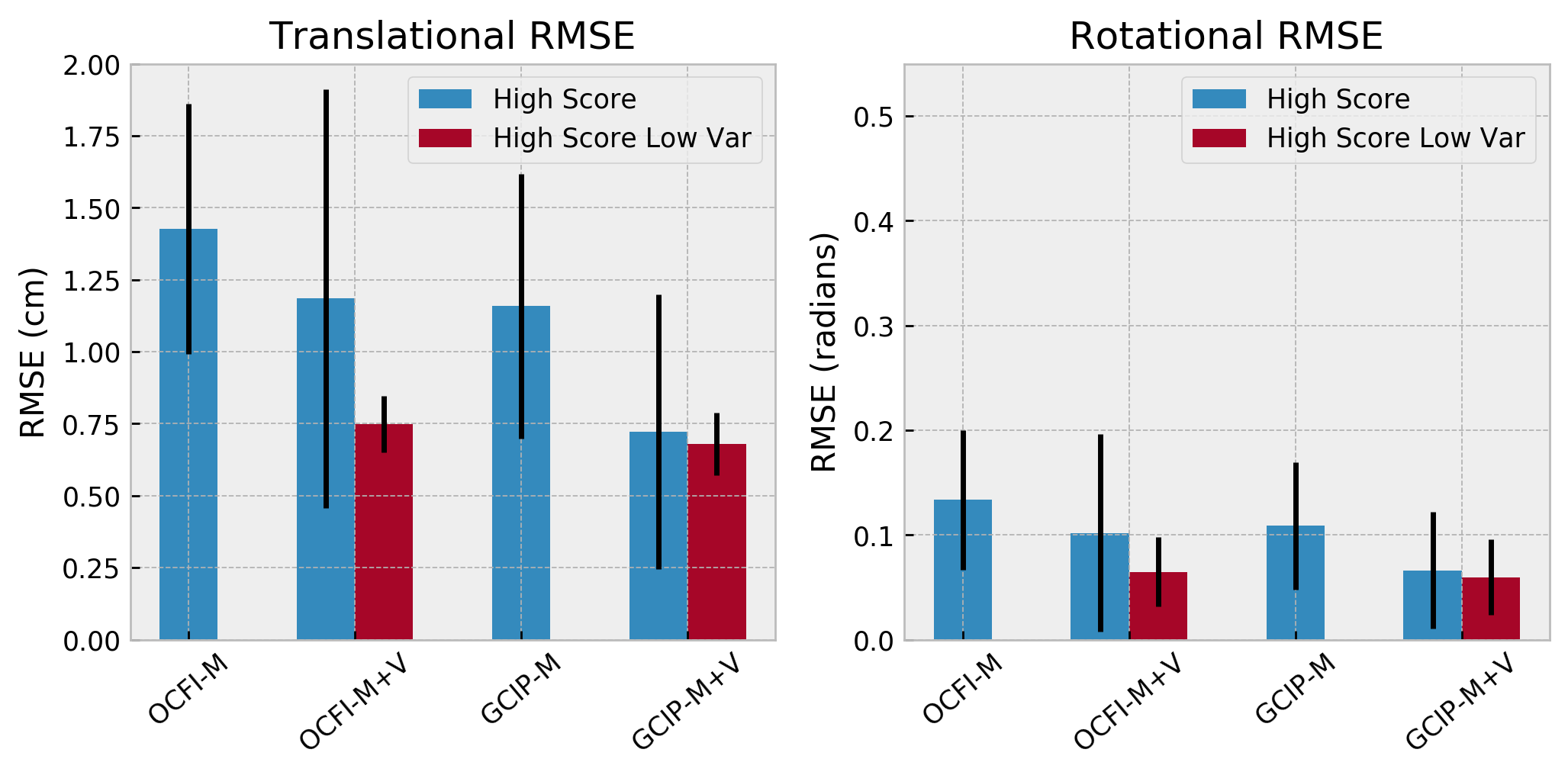}
    \caption{\textbf{Real World Translational and Rotational RMSE of Grasp Displacement Predictions.}}
    \vspace{-5mm}
    \label{fig:robot_test}
\end{figure}

We carried out the same two displacement estimation experiments with real world robot: (1) evaluate on high quality grasps and (2) evaluate on high quality and low-variance grasps.
Because we do not have a set of successful training grasps for these objects, \textbf{LOWESS} could not be applied.
Results are shown in Fig.\ref{fig:robot_test}. 
In the first experiments, by choosing highest GQN scored grasp in each trial, the robot achieved an \textbf{86.3\%} lifting success rate. 

The \textbf{GCIP-M+V} GDN has the best displacement estimation in both experiments. 
In the first experiment it achieved a translational RMSE of \textbf{0.72cm} and rotational RMSE of \textbf{3.79deg}. 
These errors are further reduced to \textbf{0.68cm} and \textbf{3.42deg} respectively by choosing high quality grasps with low predicted displacement variance. 
Although the mean RMSE values are similar across the two experiments, the error bars are greatly decreased when using the high-score low-variance grasps, which indicates that these models are more consistent and robust across different objects.

In Figure~\ref{fig:robot_test} we show an instance of the palletizing application, wherein we use \textbf{GCIP-M+V} to estimate the post-grasp displacement and compensate for the placing action accordingly.

\begin{figure}[!ht]
    \centering
    \includegraphics[width=\textwidth]{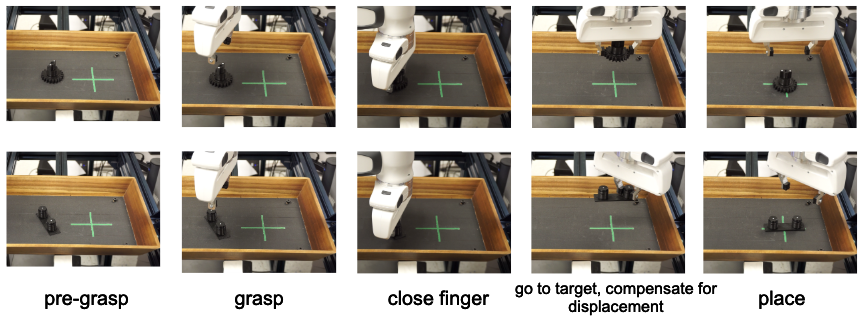}
    \caption{\textbf{Example Palletizing Application.} We use the post-grasp displacement prediction from our \textbf{GCIP-M+V} model to accurately place test objects at a target pose. The target location is marked by the center of the green cross and the target orientation is the horizontal axis of the green cross. By compensating the placing target pose with the estimated post-grasp displacement, the objects can be placed more accurately.}
    \label{fig:robot_exp}
\end{figure}
\vspace{-1em}
\section{Conclusion}
In this work we propose a method to plan robust and precise grasps by training two networks - one to predict grasp robustness and the other to predict the distribution of post-grasp object displacements.
We trained the networks in simulation and deployed them in the real world without further fine-tuning.
Experiments in simulation and the real world show our method can effectively predict post-grasp displacements, and choosing grasps that have low predicted variance result in lower displacement prediction errors.
\label{sec:6}

\begin{acknowledgement}
This project was funded and supported by Epson.
This project is also in part supported by the National Science Foundation Graduate Research Fellowship Program under Grant No. DGE 1745016. 
The authors also thank Kevin Zhang for his help with real world robot experiments.
\end{acknowledgement}

\bibliographystyle{spbasic}
\bibliography{main}

\end{document}